%% file: main.tex
\pdfoutput=1
\documentclass{article}
\usepackage{arxiv_article,times}
\usepackage{amsmath,amssymb,booktabs,graphicx,xcolor,hyperref,url,multirow}
\usepackage{tikz}\usetikzlibrary{arrows.meta,positioning,calc}
\usepackage{algorithm,algpseudocode}
\usepackage{enumitem}
\usepackage{longtable}
\usepackage{wrapfig}
\hypersetup{pdftitle={Imperfect Is Perfect: Learned Column Prefilling for Branch-Price-and-Cut},pdfauthor={Haoran Liu and Yu Yang},hypertexnames=false}
\AtBeginDocument{\setlength{\abovedisplayskip}{5pt plus 1pt minus 1pt}\setlength{\belowdisplayskip}{5pt plus 1pt minus 1pt}\setlength{\abovedisplayshortskip}{2pt}\setlength{\belowdisplayshortskip}{3pt}}

\input{numbers}

\input{numbers_writing}
\title{Imperfect Is Perfect: Learned Column\\Prefilling for \mbox{Branch-Price-and-Cut}}
\author{Haoran Liu\\
Department of Industrial Systems Engineering\\
University of Florida\\
Florida, United States\\
\texttt{haoran.liu@ufl.edu}
\And
Yu Yang\\
Department of Industrial Systems Engineering\\
University of Florida\\
Florida, United States\\
\texttt{yu.yang@ise.ufl.edu}
}
\begin{document}
\maketitle
\begin{abstract}
\input{sections/abstract}
\end{abstract}
\input{sections/intro}
\input{sections/background}
\input{sections/method}
\input{sections/experiments}
\input{sections/conclusion}

\bibliography{refs}
\bibliographystyle{arxiv_article}
\appendix
\input{sections/appendix}
\end{document}

%% file: numbers.tex
\newcommand{\archiveFastPathLo}{24}
\newcommand{\archiveFastPathHi}{52}

\newcommand{\ablVfourMainInstN}{95}
\newcommand{\ablVfourMainPnGM}{0.261}

\newcommand{\ablVfourMainConlyGM}{0.616}

\newcommand{\ablVfourMainGGM}{0.884}

\newcommand{\ablVfourMainNcpuGM}{0.264}

\newcommand{\ablVfourMainGpuOnStackPooledGM}{0.500}

\newcommand{\ablVfourMainGpuOnStackPooledN}{95}

\newcommand{\nhlRtxVfourMain}{0.132}
\newcommand{\nhlRtxVfourMainSpeed}{7.6}

\newcommand{\nhlRtxVfourMainBest}{0.057}

\newcommand{\gpuHwLFourOverCpuSpeedMed}{4.70}

\newcommand{\gpuHwRtxOverCpuSpeedMed}{8.90}

\newcommand{\gpuHwLargestInst}{X-n1001-k43}
\newcommand{\gpuHwLargestCpuSec}{15.84}

\newcommand{\gpuHwLargestLFourSec}{4.02}

\newcommand{\gpuHwLargestBTwoSec}{1.21}

\newcommand{\gpuHwLargestRtxSec}{1.30}

%% file: numbers_writing.tex
\newcommand{\upmsBodyHMFourNForty}{3.1 (5)}
\newcommand{\upmsBodyHMFourNHundred}{1800$^{*}$ (0)}
\newcommand{\upmsBodyHMEightNForty}{1.4 (5)}
\newcommand{\upmsBodyHMTwelveNHundred}{403.4 (5)}
\newcommand{\upmsBodyHMTwentyNHundred}{197.8 (5)}
\newcommand{\upmsBodyHAll}{57.1 (20)}
\newcommand{\upmsBodyHnnMFourNForty}{17.7 (5)}
\newcommand{\upmsBodyHnnMFourNHundred}{1800$^{*}$ (0)}
\newcommand{\upmsBodyHnnMEightNForty}{12.2 (5)}
\newcommand{\upmsBodyHnnMTwelveNHundred}{866.4 (5)}
\newcommand{\upmsBodyHnnMTwentyNHundred}{434.6 (5)}
\newcommand{\upmsBodyHnnAll}{171.2 (20)}
\newcommand{\upmsBodyHPMFourNForty}{1.3 (5)}
\newcommand{\upmsBodyHPMFourNHundred}{1800$^{*}$ (0)}
\newcommand{\upmsBodyHPMEightNForty}{0.2 (5)}
\newcommand{\upmsBodyHPMTwelveNHundred}{6.6 (5)}
\newcommand{\upmsBodyHPMTwentyNHundred}{1.8 (5)}
\newcommand{\upmsBodyHPAll}{5.4 (20)}
\newcommand{\upmsBodyHwMFourNForty}{1.6 (5)}
\newcommand{\upmsBodyHwMFourNHundred}{1489$^{*}$ (2)}
\newcommand{\upmsBodyHwMEightNForty}{1.1 (5)}
\newcommand{\upmsBodyHwMTwelveNHundred}{275.7 (5)}
\newcommand{\upmsBodyHwMTwentyNHundred}{154.1 (5)}
\newcommand{\upmsBodyHwAll}{40.9 (22)}
\newcommand{\upmsBodyHwPMFourNForty}{0.6 (5)}
\newcommand{\upmsBodyHwPMFourNHundred}{1047$^{*}$ (4)}
\newcommand{\upmsBodyHwPMEightNForty}{0.2 (5)}
\newcommand{\upmsBodyHwPMTwelveNHundred}{7.6 (5)}
\newcommand{\upmsBodyHwPMTwentyNHundred}{2.9 (5)}
\newcommand{\upmsBodyHwPAll}{5.1 (24)}
\newcommand{\upmsBodyHwPCMFourNForty}{0.7 (5)}
\newcommand{\upmsBodyHwPCMFourNHundred}{628 (5)}
\newcommand{\upmsBodyHwPCMEightNForty}{0.2 (5)}
\newcommand{\upmsBodyHwPCMTwelveNHundred}{7.6 (5)}
\newcommand{\upmsBodyHwPCMTwentyNHundred}{3.3 (5)}
\newcommand{\upmsBodyHwPCAll}{4.8 (25)}
\newcommand{\ablTableVfourMainRowsWords}{%
X & short & 5 & 0.149 (5/5) & 0.686 (5/5) & 1.002 (3/5) & 0.193 (5/5) & 0.125 (5/5) \\
X & medium & 5 & 0.154 (5/5) & 0.671 (5/5) & 0.925 (5/5) & 0.173 (5/5) & 0.100 (5/5) \\
X & long & 5 & 0.269 (5/5) & 0.587 (5/5) & 0.886 (5/5) & 0.258 (5/5) & 0.101 (5/5) \\
X & very long & 5 & 0.318 (5/5) & 0.689 (5/5) & 0.870 (5/5) & 0.292 (5/5) & 0.133 (5/5) \\
\cmidrule(lr){1-8}
\textbf{X} & all & 20 & 0.210 (20/20) & 0.657 (20/20) & 0.919 (18/20) & 0.224 (20/20) & 0.114 (20/20) \\
\midrule
XML300 & short & 5 & 0.203 (5/5) & 0.710 (5/5) & 0.969 (5/5) & 0.234 (5/5) & 0.136 (5/5) \\
XML300 & medium & 5 & 0.282 (5/5) & 0.677 (5/5) & 0.939 (5/5) & 0.310 (5/5) & 0.141 (5/5) \\
XML300 & long & 5 & 0.270 (5/5) & 0.690 (5/5) & 0.954 (4/5) & 0.274 (5/5) & 0.179 (5/5) \\
XML300 & very long & 5 & 0.452 (5/5) & 0.642 (5/5) & 0.865 (5/5) & 0.421 (5/5) & 0.208 (5/5) \\
XML300 & ultra long & 5 & 0.392 (5/5) & 0.329 (5/5) & 0.742 (4/5) & 0.296 (5/5) & 0.150 (5/5) \\
\cmidrule(lr){1-8}
\textbf{XML300} & all & 25 & 0.307 (25/25) & 0.588 (25/25) & 0.890 (23/25) & 0.301 (25/25) & 0.161 (25/25) \\
\midrule
XML400 & short & 5 & 0.179 (5/5) & 0.717 (5/5) & 0.955 (5/5) & 0.209 (5/5) & 0.117 (5/5) \\
XML400 & medium & 5 & 0.237 (5/5) & 0.660 (5/5) & 0.931 (4/5) & 0.238 (5/5) & 0.115 (5/5) \\
XML400 & long & 5 & 0.300 (5/5) & 0.668 (5/5) & 0.932 (4/5) & 0.275 (5/5) & 0.131 (5/5) \\
XML400 & very long & 5 & 0.369 (5/5) & 0.660 (4/5) & 0.848 (5/5) & 0.373 (4/5) & 0.170 (5/5) \\
XML400 & ultra long & 5 & 0.465 (5/5) & 0.493 (5/5) & 0.734 (5/5) & 0.391 (5/5) & 0.169 (5/5) \\
\cmidrule(lr){1-8}
\textbf{XML400} & all & 25 & 0.293 (25/25) & 0.634 (24/25) & 0.876 (23/25) & 0.288 (24/25) & 0.138 (25/25) \\
\midrule
XML500 & short & 5 & 0.153 (5/5) & 0.739 (5/5) & 0.941 (5/5) & 0.182 (5/5) & 0.082 (5/5) \\
XML500 & medium & 5 & 0.165 (5/5) & 0.648 (5/5) & 0.938 (4/5) & 0.168 (5/5) & 0.103 (5/5) \\
XML500 & long & 5 & 0.231 (5/5) & 0.668 (5/5) & 0.955 (5/5) & 0.236 (5/5) & 0.104 (5/5) \\
XML500 & very long & 5 & 0.336 (5/5) & 0.637 (5/5) & 0.902 (5/5) & 0.395 (5/5) & 0.159 (5/5) \\
XML500 & ultra long & 5 & 0.368 (5/5) & 0.371 (5/5) & 0.616 (5/5) & 0.283 (5/5) & 0.148 (5/5) \\
\cmidrule(lr){1-8}
\textbf{XML500} & all & 25 & 0.235 (25/25) & 0.597 (25/25) & 0.859 (24/25) & 0.241 (25/25) & 0.116 (25/25) \\
\midrule
\textbf{all} & & 95 & 0.261 (95/95) & 0.616 (94/95) & 0.884 (88/95) & 0.264 (94/95) & 0.132 (95/95) \\}
\newcommand{\provInst}{X-n1001-k43}

\newcommand{\nhlRtxVfourMainBestPct}{\fpeval{round(\nhlRtxVfourMainBest*100,1)}}
\newcommand{\nhlRtxVfourMainPct}{\fpeval{round(\nhlRtxVfourMain*100,1)}}
\newcommand{\nhlRtxVfourMainBestSpeed}{\fpeval{round(1/\nhlRtxVfourMainBest,1)}}
\newcommand{\ablVfourMainPnPct}{\fpeval{round(\ablVfourMainPnGM*100,1)}}
\newcommand{\upmsAllN}{25}
\newcommand{\upmsBodyHAllGM}{57.1}
\newcommand{\upmsBodyHAllCert}{20}
\newcommand{\upmsBodyHwPCAllGM}{4.8}

\newcommand{\upmsHOverHwPCAllSpeed}{\fpeval{round(\upmsBodyHAllGM/\upmsBodyHwPCAllGM,0)}}
\newcommand{\ablVfourMainConlyPct}{\fpeval{round(\ablVfourMainConlyGM*100,1)}}

\newcommand{\shareVanExact}{15}

\newcommand{\shareNcpuExact}{61}

%% file: sections/abstract.tex
Column generation folklore warns that initial columns hurt: a warm start may bias the dual and slow
the root of branch-price-and-cut. We show the opposite once the columns are imperfect in the right
way. Preloading the support of the optimal root LP, the best columns one could know in advance, fails to
finish on half of our hardest instances; a pool of ordinary routes that need not belong to any optimal
solution, held outside the master and priced at the current smoothed dual, speeds the root up. Because the columns need not be optimal, producing them becomes a learning task: a
neural edge scorer guides a constructive generator that builds the pool in seconds. A second
network predicts the initial center of Wentges smoothing from the instance file, and with it the
direction of the smoothed pricing queries; the solver's own updates run unchanged. With prefill and center
in place, exact pricing dominates root pricing time, so we rebuild bucket-graph labeling around its
dominance test for the GPU and certify on the CPU. Inside RouteOpt, the state-of-the-art exact
solver for capacitated vehicle routing, the pipeline reaches the same certified root bound in
\nhlRtxVfourMainPct{}\,\% of vanilla RouteOpt's root pricing time over \ablVfourMainInstN{} instances with
$300$ to $1000$ customers, prefill alone in \ablVfourMainPnPct{}\,\%, and \nhlRtxVfourMainBestSpeed{}
times faster in the best case. In an exact branch-price-and-cut solver for unrelated parallel
machine scheduling, prefill with
smoothing and the learned center certifies all \upmsAllN{} benchmark instances,
\upmsHOverHwPCAllSpeed{} times faster than the solver's own column generation, which certifies \upmsBodyHAllCert{} of them.

%% file: sections/intro.tex
\section{Introduction}
\label{sec:intro}

Branch-price-and-cut (BPC) is the leading exact approach for large-scale NP-hard
optimization problems, such as vehicle routing, crew and machine scheduling, and cutting stock \citep{barnhart1998branch, lubbecke2005selected,
vandenakker1999parallel, costa2019exact}. Within branch-and-bound, column generation (CG) solves the
Dantzig--Wolfe relaxation by generating necessary variables on the fly through solving a pricing subproblem, and
cutting planes are typically applied to strengthen the bound. Work on accelerating BPC has therefore concentrated on
branching, pricing, and cutting \citep{fukasawa2006robust, jepsen2008subset, baldacci2011new,
pecin2017improved, pessoa2020generic, sadykov2021bucket}, with arc elimination
\citep{irnich2010path}, route enumeration \citep{baldacci2008exact} and dual stabilization
\citep{dumerle1999stabilized, wentges1997weighted, pessoa2018automation} as complementary
accelerators. All of these techniques apply across problems, but pricing is where the problem's structure
enters. Our main test bed is RouteOpt \citep{routeopt}, an exact solver for the capacitated
vehicle routing problem (CVRP) with a state-of-the-art integrated implementation of all of the above.

With the CPU-side toolbox this mature, learning and GPU computing offer two further directions. Learning has
been applied to branching \citep{khalil2016learning, gasse2019exact, bengio2021machine, you2026two}, to cutting plane generation and
selection \citep{tang2020reinforcement, paulus2022learning, wang2023learning, deza2023machine}, to
column generation and selection \citep{morabit2021machine, chi2022deep, morabit2023arc,
shen2022enhancing, xu2025enhancing, koutecka2025ranking, vaclavik2018accelerating} and to dual
stabilization \citep{kraul2023machine, babaki2022coil, shen2024adaptive, sarkar2025accelerating,
fang2025learning, wang2026learning}. On the GPU side, first-order methods have moved the linear-programming solve onto the GPU
\citep{lu2025cupdlp}, and GPU labeling exists for constrained shortest paths outside BPC
\citep{lu2021accelerating}, but the exact pricer of a state-of-the-art BPC solver has stayed on the CPU. It is worth noting that gains beyond a modest research
baseline can shrink or reverse inside a state-of-the-art solver, so a method of practical value has
to be demonstrated against an efficient implementation.

Our target is \emph{root pricing}: the column generation that solves the root relaxation before
any cutting plane is added. Its wall-clock time is our outcome, and every configuration we compare that completes ends at
the same certified root bound, so root pricing is a clean test bed and delivers a certified lower
bound of practical value on its own. Our method is a combination of three moves: a prefilled
archive supplies candidate columns, a learned initial center shapes the pricing queries, and a GPU
accelerates the exact pricing level.

We started from a piece of folklore. Textbooks warn that a warm start, even from an excellent primal solution, can
hinder column generation \citep{lubbecke2005selected, vanderbeck2005implementing}: the initial
columns pull the dual toward the solution they encode, and in the root of a modern BPC, heuristic routes are elementary and need not suffice to attain or certify the ng-route
bound that pricing works over \citep{baldacci2011new}, while a
large pool adds to the complexity of solving the restricted master problem (RMP). Our measurements confirmed the warning on deep instances: preloading a heuristic pool into the
master slowed the root down there, and preloading the support of the optimal root LP, the best
columns one could know in advance, was the worst choice of all (Section~\ref{sec:m-prefill},
Appendix~\ref{app:earlier}). What did help was the opposite of what
the folklore protects against. A diverse pool of ordinary, non-optimal routes, held
\emph{outside} the master as a lazy archive and queried at the current smoothed dual, shortens
root pricing by a large factor on every benchmark set and route-size class we test
(Section~\ref{sec:exp-ablation}). Learning for column generation has so far concentrated on the
pricing loop itself, on selecting, ranking, or generating columns during the iterations. We take a
different perspective: leave the pricer untouched and learn the prefilling, with a constructive
generator guided by a neural edge scorer, whose pool need not be optimal.

Prefilling alone is not the whole story. In an earlier development experiment, dumping a
metaheuristic pool into the initial master slowed the root down, as the folklore predicts
(Appendix~\ref{app:results-pool}): the pricer spent many rounds searching for the direction of the
dual and accumulated a bloated LP. The
classical remedy is dual smoothing \citep{dumerle1999stabilized, wentges1997weighted,
pessoa2018automation}, and the learned variants predict the optimal dual and install it as the
reference point \citep{kraul2023machine, shen2024adaptive}. Where those methods learn the values of the dual, we learn its direction: the RMP has many optimal duals, so what
matters is the direction from which the query approaches the optimum, not its distance to one
optimal point. We therefore learn the \emph{initial} center of the standard Wentges smoothing from the
instance file, calibrate its scale, and let the solver's own rule move the center and adapt the
weight from there. On the hardest instances in particular, the learned center becomes decisive, especially in combination with prefill (Section~\ref{sec:exp-ablation}, Appendix~\ref{app:results-center}).

Prefill and center cut the run time of heuristic pricing sharply, so exact pricing, the level that certifies
optimality, becomes the dominant share of root time. Bucket-graph labeling is the exact pricer of
modern routing solvers \citep{sadykov2021bucket}, and its dominance
test is the cost that structure exists to contain; parallel CPU labeling has been developed for it
\citep{petersen2025parallel}, while GPU work on BPC has addressed the linear-programming solve \citep{lu2025cupdlp}. We rebuild exact bucket-graph labeling around the
dominance test for the GPU, with a final CPU certificate
(Sections~\ref{sec:m-gpu} and \ref{sec:exp-gpu}). The construction applies to any bucket-graph-based pricer once the
resource extension, the dominance rule, and the join are adapted to the problem. Our
\textbf{contributions} are summarized as follows.
\begin{itemize}[nosep,leftmargin=*]
\item A general learn-to-prefill framework: a neural edge scorer and a constructive generator whose pool
      is served to root pricing as a lazy archive. This opens a new direction for learning in column
      generation, one that is both highly effective and broadly applicable (Section~\ref{sec:m-prefill};
      Appendix~\ref{app:earlier}).
\item A learned dual direction under prefilled columns: the initial center of Wentges smoothing,
      predicted by a neural network, together with a calibration that emphasizes the direction and
      relaxes its absolute magnitude to better steer the prefilled columns (Section~\ref{sec:m-centre},
      Appendix~\ref{app:results-centre}).
\item A GPU backend for exact bucket-graph labeling with CPU certification, which applies to
      any BPC solver with bucket-graph pricing (Sections~\ref{sec:m-gpu} and \ref{sec:exp-gpu}).
\item A general plug-in pipeline for problems solved by BPC, which delivers strong performance
      inside optimized exact BPC solvers: the state-of-the-art CVRP solver RouteOpt and an exact
      solver for unrelated parallel machine scheduling (Sections~\ref{sec:exp-ablation} to
      \ref{sec:exp-upms}).
\end{itemize}

%% file: sections/background.tex
\section{Background}
\label{sec:background}

\paragraph{Set partitioning, its dual, and pricing.}
\label{sec:bg-master}
Let $C=\{1,\dots,n\}$ be the customers, $0$ the depot, $q_i$ the demand of customer $i$, $Q$ the
capacity, and $d_{0i}$ the depot distance. Over a route set $\mathcal{R}$, the master problem and its
dual are
\begin{equation}
\min_{\lambda\ge 0}\Big\{\sum_{r\in\mathcal{R}} c_r\lambda_r:\ \sum_{r\in\mathcal{R}} a_{ir}\lambda_r=1,\ i\in C\Big\}
\;=\;
\max_{u\in\mathbb{R}^{n}}\Big\{\sum_{i\in C} u_i:\ rc_r(u)\ge 0,\ r\in\mathcal{R}\Big\},
\label{eq:master}
\end{equation}
with $c_r$ the cost of route $r$, $a_{ir}$ its number of visits to customer $i$ and
$rc_r(u)=c_r-\sum_{i\in C}a_{ir}u_i$ its reduced cost. A dual point $u$ is feasible exactly when
$\min_r rc_r(u)\ge 0$, so \emph{pricing} is the separation problem of the dual. Column generation
solves the RMP over a subset of $\mathcal{R}$, starting from the
singleton routes $(0,i,0)$, prices at its dual $u$ and adds the routes found
\citep{barnhart1998branch, lubbecke2005selected}; when pricing certifies $u$, the RMP value is the
root bound. Elementary pricing is intractable at these sizes; the ng-route relaxation
\citep{baldacci2011new}, which lets a route revisit a customer only after visiting a customer whose neighborhood does
not contain it, made the exact solution of large instances practical, and $\mathcal{R}$ is the set of
ng-feasible routes in RouteOpt as in other modern solvers (Appendix~\ref{app:bpc-cvrp}). This paper calls \emph{prefilling} any supply of candidate columns prepared before column
generation starts, whether preloaded into the initial RMP or held outside it and served on demand.

\paragraph{Heuristic and exact pricing.}
\label{sec:bg-levels}
Pricing is a shortest-path problem with resource constraints, solved by bucket-graph labeling
\citep{sadykov2021bucket} at three levels, as in generic BPC solvers \citep{pessoa2020generic}:
two \emph{heuristic} levels (LIGHT and HEAVY) drop part of the label state, are cheap, may miss
columns and account for most of the iterations; the \emph{exact} level keeps every state the
dominance rule needs, is expensive, and alone certifies that no column prices out. Root pricing stops when an exact pricing call at the raw dual $u$ that explores its whole label space, a \emph{complete} call, returns no route with
$rc_r(u)<-\epsilon$, where $\epsilon>0$ is the solver's pricing tolerance; its wall clock is what we measure (Section~\ref{sec:exp-setup-main}). This
pricer is highly efficient, a bar that learned column generators rarely clear, which is why our two learned
components leave its algorithm untouched; the GPU component of Section~\ref{sec:m-gpu} reimplements
its exact level without changing what it returns.

\paragraph{Dual stabilization.}
\label{sec:bg-smoothing}
The master \eqref{eq:master} is highly degenerate, so the RMP dual jumps between optimal solutions
from one iteration to the next \citep{dumerle1999stabilized, pessoa2018automation}. Wentges
smoothing \citep{wentges1997weighted} prices at a convex combination of the RMP dual $u_t$ and a
stabilization center $c_t$,
\begin{equation}
q_t\ =\ (1-\alpha_t)\,u_t+\alpha_t\,c_t ,\qquad \alpha_t\in[0,1),
\label{eq:smooth}
\end{equation}
where $c_t$ is the dual point with the best certified Lagrangian bound so far and $\alpha$ is
adapted from $\alpha_0=0.5$ by the solver's weight rule, \eqref{eq:alpha} below; candidates
are re-priced at $u_t$ and admitted only if $rc_r(u_t)<-\epsilon$, so the termination test is
unchanged.

%% file: sections/method.tex
\section{Method}
\label{sec:method}

\begin{figure}[!tb]
\centering
\begin{tikzpicture}[x=1cm,y=1cm,
  box/.style={draw, rounded corners=1.5pt, align=center, inner xsep=3pt, inner ysep=1pt,
              font=\scriptsize, minimum height=8.5mm},
  nn/.style={box, fill=black!8},
  hub/.style={draw, align=center, inner xsep=3pt, inner ysep=1pt, font=\scriptsize,
              minimum width=1.70cm, minimum height=7mm},
  arr/.style={-{Latex[length=1.5mm,width=1.1mm]}, semithick},
  pipe/.style={semithick},
  lab/.style={font=\tiny, inner sep=1.2pt, align=center}]

\node[hub] (rmpc) at (7.00, 0.35) {\textbf{RMP}\\[-1.5pt]{\tiny columns so far}};
\node[hub] (lpc)  at (7.00,-0.35) {LP solver\\[-1.5pt]{\tiny primal simplex}};

\node[box, minimum width=1.05cm] (inst)   at (0.550, 0.35) {instance\\file};
\node[nn,  minimum width=1.85cm] (scorer) at (3.075, 1.60)
      {\textbf{edge scorer}\\[-1.5pt]{\tiny message passing}\\[-1.5pt]{\tiny on the candidate graph}};
\node[box, minimum width=1.45cm] (gen)  at (5.225,1.60) {constructive\\generator};
\node[box, minimum width=1.30cm] (pool) at (7.000,1.60) {route pool\\[-1.5pt]{\tiny lazy archive}};
\node[box, minimum width=1.35cm] (cert) at (11.000,1.60)
      {certified\\root bound\\[-1.5pt]{\tiny CPU exact call}};

\node[nn,  minimum width=2.00cm] (st) at (2.650,-1.60)
      {\textbf{set transformer}\\[-1.5pt]{\tiny customer tokens,}\\[-1.5pt]{\tiny distance-biased attention}};
\node[box, minimum width=1.85cm] (c0) at (5.200,-1.60)
      {initial center $c_0$\\[-1.5pt]{\tiny direction and scale}};
\node[draw, circle, inner sep=0pt, minimum size=4.6mm, font=\scriptsize] (mix) at (7.000,-1.60) {$\oplus$};
\node[box, minimum width=1.95cm] (heur) at (10.400,-1.60)
      {heuristic labeling\\[-1.5pt]{\tiny LIGHT, then HEAVY}};
\node[box, minimum width=1.65cm] (gpu)  at (11.000, 0.35) {exact labeling\\on the \textbf{GPU}};

\draw[arr] (inst.east) -- ++(0.12,0) |- (scorer.west);
\draw[arr] (inst.east) -- ++(0.12,0) |- (st.west);
\draw[arr] (inst.east) -- node[lab,above] {singleton columns} (rmpc.west);

\draw[arr] (scorer.east) -- (gen.west);
\draw[arr] (gen.east) -- (pool.west);
\draw[arr] (pool.south) -- (rmpc.north);
\node[lab,anchor=west] at (7.10,0.94) {routes priced out at $q_t$, admitted at $u_t$};

\draw[arr] (st.east) -- (c0.west);
\draw[arr] (lpc.south) -- (mix.north);
\node[lab,anchor=west] at (7.10,-1.03) {dual $u_t$};
\draw[arr] (c0.east) -- node[lab,above] {$c_t$} (mix.west);
\node[lab,anchor=north] at (7.00,-2.12) {smoothed query $q_t=(1-\alpha_t)\,u_t+\alpha_t\,c_t$};

\draw[arr] (mix.east) -- node[lab,above] {$q_t$} (heur.west);
\draw[arr] ([xshift=5mm]heur.north) -- node[lab,right,pos=0.62] {no column} ([xshift=-1mm]gpu.south);
\draw[arr] (gpu.north) -- node[lab,anchor=west] {} (cert.south);
\node[lab,anchor=west] at (11.15,0.975) {no column};

\draw[pipe] ([xshift=-5mm]heur.north) -- (9.900,-0.95) -- (8.600,-0.95) -- (8.600,0.35);
\draw[pipe] (gpu.west) -- (8.600,0.35);
\draw[arr]  (8.600,0.35) -- (rmpc.east);
\fill (8.600,0.35) circle (0.55pt);
\node[lab,anchor=south] at (9.80,-0.92) {$rc_r(u_t)<-\epsilon$};
\end{tikzpicture}
\caption{The pipeline. Two neural models act on the primal and the dual side of the RMP, the LP solver connects the two, and exact pricing runs on the GPU with CPU certification.}
\label{fig:pipeline}
\end{figure}
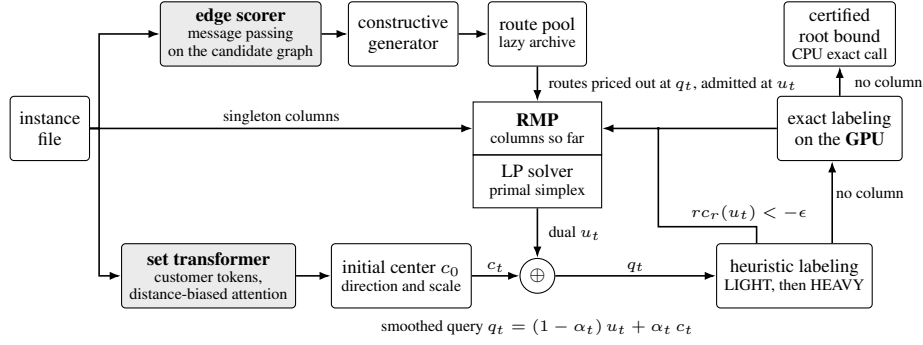

Our method combines three moves, arranged as in Figure~\ref{fig:pipeline}. Before the run, two
neural models read the instance file. An \emph{edge scorer} drives a constructive generator that
builds a large pool of candidate routes, and a \emph{set transformer} predicts the initial center
of dual smoothing. Root pricing then starts from the singleton columns and alternates two loops.
In the first, the smoothed query $q_t$ of \eqref{eq:smooth} is served first from the pool, which
acts as a lazy archive, and then by heuristic pricing, with a primal simplex re-solve of the RMP
after each batch. When the heuristic levels return nothing, exact pricing takes over on the GPU, again alternating
with primal simplex, until a complete exact call at the raw dual, re-proved on the CPU, returns
nothing and the root bound is certified.

\subsection{Learning to prefill}
\label{sec:m-prefill}\label{sec:m-archive}

\paragraph{Columns act on the dual.}
A set $T$ of columns in the master restricts the dual of \eqref{eq:master} to the polyhedron
\begin{equation}
D_T=\{u:\ a_r^{\top}u\le c_r,\ r\in T\}\ \supseteq\ D_{\mathcal{R}},\qquad
F_T=\arg\max\{\textstyle\sum_i u_i:\ u\in D_T\},
\label{eq:dualpoly}
\end{equation}
where $a_r$ is the visit vector of route $r$, for an elementary route the incidence vector of its
customer set; RouteOpt's vehicle-count row is suppressed here and restored in \eqref{eq:lagr}
(Appendix~\ref{app:bpc-cg} states $D_T$ and $F_T$ with it). The singleton columns make the RMP
feasible and bound its dual objective above, and primal simplex returns a basic optimal dual $u_T\in F_T$, a
vertex of $D_T$. In exact arithmetic root pricing terminates exactly when
$u_T\in D_{\mathcal{R}}$, that is, when $m(u_T)=\min_{r}rc_r(u_T)\ge 0$; the implementation
uses the pricing tolerance of Section~\ref{sec:background}. This is why preloading the support of an optimal solution failed in our hands (Appendix~\ref{app:earlier}). If $T$ contains the support of a
full-master optimum, the RMP already attains the optimal value $z^{*}$, yet
$F_T$ only meets $D_{\mathcal{R}}$: recovering the optimal primal value says nothing about
whether the vertex the simplex returns lies in $D_{\mathcal{R}}$ (Proposition~\ref{prop:app-opt}), so preloading a converged
support leaves termination to that choice. A column restricts the dual only through its visit
vector and its cost, so our design hypothesis is that what shortens the run is \emph{coverage}: many distinct customer
sets, each customer in many of them, so that \eqref{eq:dualpoly} constrains $u_T$ from many
directions along the whole trajectory; Appendix~\ref{app:results-pool} evaluates this design against
preloading. A seed-based population heuristic produces
exactly such coverage, its seeds exploring different regions of the route space; hybrid genetic
search \citep{vidal2022hybrid} is the strongest known metaheuristic for the CVRP, and we learn to reproduce
the routes it samples.

\paragraph{Learning task.}
Let $E(I)$ be the candidate edge set of instance $I$, a nearest-neighbor graph plus the depot
edges. Among the reference-pool routes that price out at the reference center, the tree-predicted dual of
Appendix~\ref{app:learn-prefill-train}, let $\nu_e\in[0,1]$ be the fraction of those visiting $i$
or $j$ that traverse the edge $e=(i,j)$; these are the \emph{heated} edges, which measure edge usage in the routes the reference dual
asks for. The scorer $f_\theta$ maps the instance-only node and edge features $\Phi(I)$
(Appendix~\ref{app:learn-prefill-train}) to a heat map on $E(I)$,
\begin{equation}
\hat z=f_\theta(\Phi(I)),\qquad \hat\nu_e=\big[\max\{\hat z_e,0\}\big]^2,\qquad
\mathcal{L}_{\mathrm{edge}}(\theta)=\sum_{I}\ \sum_{e\in H(I)}\big(\hat z_e-\sqrt{\nu_e}\big)^2 ,
\label{eq:heat}
\end{equation}
where the supervised set $H(I)$ adds the reference-pool edges to $E(I)$; message passing runs on
$E(I)$ and the edge head is evaluated on $H(I)$ in training and on $E(I)$ at generation, and the
model is selected on validation rank correlation with $\nu$. Heat is contextual, decided by an edge's
competitors at both endpoints, so $f_\theta$ is a message-passing network with messages symmetric
in the endpoints. Its graph is constructed from the instance alone and is identical during
training and generation.

\paragraph{From heat map to columns.}
Algorithm~\ref{alg:gen} turns the heat map into a pool $P$ of routes; its constants are listed in
Appendix~\ref{app:learn-prefill-infer}. The pool is kept outside the initial RMP and supplies columns lazily. At
pricing round $t$, before any labeling, the archive offers its $B$ most negative eligible routes (the $k_t$ largest values of $-rc_r(q_t)$) and
re-prices them at the raw dual,
\begin{align}
A_t&=\{r\in P_t:\ rc_r(q_t)<0\},\qquad
O_t=\operatorname{Top}_{k_t}\big(-rc_r(q_t);\,A_t\big),\quad k_t=\min\{B,|A_t|\},\label{eq:archive}\\
C_t&=\{r\in O_t:\ rc_r(u_t)<-\epsilon\},\qquad P_{t+1}=P_t\setminus O_t,\notag
\end{align}
so each route is offered once, and $C_t$ is admitted by the same test as a generated column, so
the archive cannot change the value CG converges to. The labeling
call is skipped when $C_t\neq\emptyset$. The recipe carries over to any problem with a
seed-based heuristic that samples diverse columns; the scheduling case study of
Section~\ref{sec:exp-upms} uses randomized list scheduling.

\begin{algorithm}[!tb]
\caption{Learned constructive prefill. $P\uplus S$ inserts the routes of $S$ into $P$ by customer set, keeping the cheaper ordering; $\kappa_i(P)$ counts the routes of $P$ through $i$; $\mathcal N(i)$ is the heat-filtered neighborhood of $i$; Polish is intra-route 2-opt; $\tau$ is the sampling exponent, $\delta,\delta'$ probability floors, $\gamma$ the coverage exponent, and $\mathrm{LS}$ relocate, swap, and 2-opt$^{*}$ (constants in Appendix~\ref{app:learn-prefill-infer}; split and local search run twice per tour).}
\label{alg:gen}
\begin{algorithmic}[1]
\footnotesize
\Require heat map $\hat\nu$ from \eqref{eq:heat}, seeds $w=1,\dots,W$, time budget $T$
\For{$w=1,\dots,W$ in parallel}
\State $P^{(w)}\leftarrow\emptyset$
\Repeat
  \State draw $\tau$ and a start $i_1$ with $\Pr[i_1=i]\propto\max\{\hat\nu_{0i},\delta\}\,\big(1+\kappa_i(P^{(w)})\big)^{-\gamma}$
  \State grow $\omega=(i_1,\dots,i_n)$ with $\Pr[i_{k+1}=j\mid i_1,\dots,i_k]\propto\big[\max\{\hat\nu_{i_kj},\delta'\}\,e^{\xi_{kj}}\big]^{\tau}$ over unvisited $j\in\mathcal N(i_k)$, $\xi$ independent noise, else the nearest unvisited customer
  \State $S\leftarrow\mathrm{Polish}\big(\mathrm{Split}(\mathrm{2opt}(\omega))\big)$;\quad $P^{(w)}\leftarrow P^{(w)}\uplus S$ \Comment{optimal split \citep{prins2004simple}}
  \State for each local-search pass: $S\leftarrow\mathrm{Polish}(\mathrm{LS}(S))$;\quad $P^{(w)}\leftarrow P^{(w)}\uplus S$
\Until{the time budget $T$ is spent, or consecutive tours add no new customer set}
\EndFor
\State \Return $P\leftarrow\biguplus_{w} P^{(w)}$
\end{algorithmic}
\end{algorithm}

\subsection{Learning the dual center}
\label{sec:m-centre}\label{sec:m-neural}

\paragraph{Not the optimal-dual task.}
Learned dual stabilization predicts the optimal dual and installs it as the reference point. We
train on the same certified labels, the optimal root dual $\pi^{*}=(\pi^{*}_C,\pi^{*}_{\mathrm{veh}})$
of each label instance (Appendix~\ref{app:learn-data}), but use the prediction to initialize the Wentges center, and we
judge it by the pricing trajectory that follows. The RMP admits many optimal duals, and the
smoothed query of \eqref{eq:smooth} needs a \emph{direction} to price from and a \emph{scale} at
which pricing stays productive, not proximity to one optimal point. By \eqref{eq:master}, a center
$c$ is dual feasible exactly when $m(c)\ge 0$. The solver certifies an installed center with one
exact pricing at $c$, whose outcome is the Lagrangian bound
\begin{equation}
L(c)=\sum_{i\in C} c_i + n\,\min\{0,\,m(c)\},
\label{eq:lagr}
\end{equation}
valid because every route covers at least one customer, so $\sum_r\lambda_r\le n$ for every
feasible $\lambda$ (Proposition~\ref{prop:app-lagr}). With RouteOpt's row $\sum_r\lambda_r\ge K_{\min}$, $K_{\min}$ a lower bound on the fleet size, the center carries a vehicle
dual $c_{\mathrm{veh}}\ge 0$, every reduced cost becomes $c_r-a_r^{\top}c_C-c_{\mathrm{veh}}$, and
the first term of \eqref{eq:lagr} becomes $\sum_i c_i+K_{\min}c_{\mathrm{veh}}$. The same call
reads off how aggressive the center is: at $m(c)\ge 0$ it returns no column, and a center at or
near feasibility returns no column at installation, so the columns must come from the calls that follow. The certified optimal dual is the
extreme case, feasible and pricing out nothing. We therefore calibrate the center to keep pricing
productive.
\paragraph{Learning task.}
The dual lives on the customers, so the model assigns one output to each customer: a set
transformer $g_\varphi$ \citep{lee2019set} over customer tokens, with distance-biased attention to represent the
interactions among customers, maps the instance-only customer and context features $Z(I)$
(Appendix~\ref{app:learn-centre-train}) to singleton-normalized coordinates $\hat y$ and a log magnitude
$\hat\ell$. With $D_0=\operatorname{diag}(2d_0)$, the direction is scored in the original dual
units and the magnitude as the ratio of the customer-dual norm to the singleton-cost norm,
\begin{align}
(\hat y,\hat\ell)&=g_\varphi(Z(I)),\qquad
\rho^{*}=\|\pi^{*}_C\|_2\,/\,\|2d_0\|_2,\notag\\
\mathcal{L}_{\mathrm{center}}(\varphi)&=\sum_I\Big[\,1-\cos\big(D_0\hat y,\pi^{*}_C\big)
+\beta\,\big(\hat\ell-\log\rho^{*}\big)^2\Big].
\label{eq:centreloss}
\end{align}
Direction and scale are produced separately: the installed center is
$c_0=(c_{0,C},c_{\mathrm{veh}})$ with
\begin{equation}
c_{0,C}=\Pi_{\mathcal{B}}\big(s\,D_0\hat y\big),\qquad
\mathcal{B}=\{c:\ 0\le c_i\le\max(2d_{0i}-c_{\mathrm{veh}},0)\},\qquad
s:\ \big\|\Pi_{\mathcal{B}}(s\,D_0\hat y)\big\|_2=T,
\label{eq:install}
\end{equation}
where $\Pi_{\mathcal{B}}$ clips to a singleton-based box, the multiplier $s$ is found by bisection
so that the customer-block norm after clipping equals the target $T$, the box corner being installed
when $T$ exceeds its norm, and $T$ is a calibrated
geometric mean of auxiliary neural magnitude estimates computed from customer features and
instance context, independently of the deployed direction. The vehicle dual $c_{\mathrm{veh}}$
comes from gradient-boosted trees.

\paragraph{Smoothing from a learned start.}
The predictor is evaluated once. At iteration $0$ the solver installs $c_0$ and certifies $L(c_0)$
by one exact call at $c_0$ whose columns are not added; this replaces the seeding exact call that
the stabilized baseline pays for its first center. From then on the rule of
Section~\ref{sec:bg-smoothing} runs unchanged, organized in exact-pricing sequences: a \emph{sequence} is a maximal run of consecutive exact calls at one raw dual $u_s$, the mis-price
ladder below together with the call that closes it. Sequence $s$
freezes the raw dual $u_s$, the center $c^{s}$, the incumbent bound $L_s$ and the starting weight
$\alpha^{s}$; let $\mathcal{Q}_s$ be its complete exact queries, $q_s^{\mathrm{best}}\in\arg\max_{q\in\mathcal{Q}_s}L(q)$,
and $\alpha_{\mathrm{last}}$ the weight of the last complete query $q_{\mathrm{last}}$. When the
sequence closes,
\begin{align}
(c^{s+1},L_{s+1})&=\begin{cases}(q_s^{\mathrm{best}},L(q_s^{\mathrm{best}})) & L(q_s^{\mathrm{best}})>L_s+\epsilon_L,\\ (c^{s},L_s) & \text{otherwise},\end{cases}
\label{eq:update}\\
\alpha^{s+1}&=\begin{cases}\max\{0,\ \alpha_{\mathrm{last}}-\eta\} & h_s>\epsilon_g,\\
\min\{\bar\alpha,\ \alpha_{\mathrm{last}}+\eta(1-\alpha_{\mathrm{last}})\} & h_s<-\epsilon_g,\\
\alpha_{\mathrm{last}} & \text{otherwise},\end{cases}
\qquad h_s=g(q_{\mathrm{last}})^{\top}(u_s-c^{s}),
\label{eq:alpha}
\end{align}
where $g$ is a supergradient of $L$ at the last complete query and the constants $\eta$, $\bar\alpha$, $\alpha_0$ and the tolerances are RouteOpt's defaults, which we do not change (Table~\ref{tab:app-smooth-constants}). The learned prediction initializes the center; every later center is
selected by certified Lagrangian bounds. Two fallback rules keep a miscalibrated center from stalling the queries. An
exact call at $q_t$ with no column admissible at $u_t$ is a mis-price rather than a certificate,
and while that remains so the query is retried along the ladder
$\alpha_j=\max\{0,\,1-(j+1)(1-\alpha^{s})\}$ until it reaches $u_t$ (Appendix~\ref{app:bpc-cg}). An empty smoothed LIGHT call
is retried once at $u_t$; if the retry admits columns the rest of that LIGHT level prices at the
raw dual, otherwise the solver advances to HEAVY. Once the deployed endgame switch fires
(Appendix~\ref{app:learn-centre-infer}), $\alpha$ is set to $0$ and adaptation stops.

\subsection{GPU exact pricing}
\label{sec:m-gpu}
The first two components change where root pricing spends its time. Prefill and a learned center reduce the work at the heuristic levels, and in our development runs this left exact pricing as the largest stage of the run and the next target for acceleration. The CPU profile of Appendix~\ref{app:gpu-bottleneck} identifies the
dominance test as the hotspot, and its per-level counters show that an exact call evaluates orders
of magnitude more dominance predicates than a heuristic call. Appendix~\ref{app:gpu-bucket} reviews
bucket-graph pricing: labels are extended bucket by bucket, each arrival is checked against the
current contents of its target bucket, and the resulting insertions and deletions create a
sequential dependence between the arrivals to a bucket. Our GPU exact pricing removes this
dependence in three steps.
\begin{enumerate}[leftmargin=*,nosep]
\item \emph{Batched dominance.} Labels enter a bucket in waves. Each wave passes three dominance
stages: stored labels against candidates, candidates against earlier candidates in a canonical
order, and the surviving candidates against the stored labels. The stages run in sequence and the
comparisons within a stage in parallel; every deletion keeps a direct dominator that survives the
wave, with conservative floating-point guards and a repair pass that restores any candidate left
without a surviving direct dominator.
\item \emph{Labeling redesigned around the test.} RouteOpt's push of one label at a time into the next bucket
becomes a level-synchronous, wave-based expansion in which the arrivals to a bucket are batched
instead of chained, the form parallel CPU labeling also takes \citep{petersen2025parallel}: resource levels are processed in order, waves repeat within a level
until its frontier is empty, and all label state is in structure-of-arrays form. Label extension,
dominance testing, and the join of half-routes run on the GPU; route reconstruction,
deduplication, and the final column selection run on the CPU, the heuristic levels stay on the
CPU, and dispatch sits inside the exact level, so the CG loop is unchanged.
\item \emph{Memory and filtering.} Shared-memory tiles reuse candidate prefixes across comparisons,
and the cross-bucket filter runs one warp
per frontier label and stops at a reduced-cost prefix bound.
\end{enumerate}
Bucket-graph labeling is the exact pricer of state-of-the-art vehicle-routing solvers, among them
VRPSolver and RouteOpt \citep{sadykov2021bucket,pessoa2020generic,routeopt}. They share the
label-comparison structure the three steps act on, whose dominance test is the cost the bucket graph
exists to contain \citep{sadykov2021bucket}, so the steps carry over once the resource extension,
the dominance rule, and the join are adapted. Appendices~\ref{app:gpu-backend} and
\ref{app:gpu-safeguards} give the implementation details and the safeguards behind the optimality guarantee.

%% file: sections/experiments.tex
\section{Experiments}
\label{sec:experiments}

We organize the empirical evaluation around three questions:
\begin{enumerate}[label=(\roman*),nosep,leftmargin=*]
\item \textbf{How the pipeline works.} Which component carries the gain, how the components
      interact, and where the prefilled columns go during column generation.
\item \textbf{What GPU pricing can add.} How much a batched exact pricer buys per call, on which
      GPU, and how that converts into root time once the cost has shifted to exact pricing.
\item \textbf{How general the recipe is.} Whether prefill and the learned center carry over to a
      different problem inside a different solver, as a case study on parallel-machine scheduling.
\end{enumerate}
Training, model selection and the constants of the pipeline are in Appendix~\ref{app:method}, the scheduling variants in Appendix~\ref{app:bpc-upms}; the ablations that isolate the two learned components are in
Appendix~\ref{app:ablation}.

\subsection{Experimental setup}
\label{sec:exp-setup-main}

All experiments ran on one workstation with an NVIDIA RTX PRO 6000 Blackwell GPU and an AMD
EPYC 9555 processor; only the per-call device comparison of Table~\ref{tab:gpu} and
Appendix~\ref{app:gpu-percall} adds an L4 and a B200. For CVRP the baseline is RouteOpt
\citep{routeopt}, the state-of-the-art exact solver, whose reach marks the frontier of instances
solved to optimality; for scheduling it is the BPC algorithm of
\citet{hijazi2026improving}, which accelerates pricing with a pointer transformer. On CVRP we focus
on the region exact solvers have not yet conquered, instances with $300$ customers and more. Route
size and instance size are the two factors that decide difficulty, so the panel is stratified by
both: on Vrp-Set-X \citep{uchoa2017new} we group instances by route size with the XML classification (Table~\ref{tab:appD-routesize}) and sample five instances with $300$ to $1000$ customers from each of the short,
medium, long and very-long classes; on XML \citep{queiroga2022xml} we sample five instances per
class from short to ultra-long at $300$, $400$ and $500$ customers; \ablVfourMainInstN{} instances
in all. For scheduling we use the standard benchmark distribution of \citet{hijazi2026improving}.
The measured quantity is the wall clock of root pricing; every completed arm reaches the same certified root bound, each configuration is
normalized to the mean of two runs of RouteOpt (\emph{vanilla}) on the same instance, and ratios are aggregated as
instance-equal geometric means.

\input{sections/tables}

\subsection{The pipeline taken apart}
\label{sec:exp-ablation}

Table~\ref{tab:ablation} compares five configurations with vanilla RouteOpt on the panel: the learned
archive alone (\emph{Prefill}), the learned center alone (\emph{Center}), the GPU backend alone on
the vanilla solver (\emph{GPU}), archive and center together, and the full pipeline. Prefill is the central
idea and the largest lever. On its own it brings root pricing to \ablVfourMainPnPct{}\,\% of vanilla's, faster on all
instances, and does so on every benchmark set and every route-size class of the panel. In instrumented runs the archive supplied columns in \archiveFastPathLo{} to \archiveFastPathHi{}\,\% of the LIGHT rounds, bypassing labeling in
those rounds, so it acts as a fast alternative to the LIGHT level; when it supplies no admissible
column, the solver resumes labeling (Appendix~\ref{app:learn-prefill-infer}). Appendix~\ref{app:results-pool} compares the lazy
archive with preloading, with a heuristic pool in place of the learned one, and with two oracles that
preload the converged column set or its support. The center is designed to correct what prefill can do to the early dual trajectory; alone it is a smaller lever, bringing root pricing to \ablVfourMainConlyPct{}\,\% of vanilla's. Prefill+Center matches Prefill on the panel average (\ablVfourMainNcpuGM{} against \ablVfourMainPnGM{}) because the center acts mainly on the hard instances, the large ones with long routes: on the ultra-long rows of every XML set it is clearly faster than Prefill alone, on long and very-long routes the two are even, and on short and medium routes, where prefill leaves little to correct, it costs a little. The controls for the center, standard smoothing and a permuted direction, are in Appendix~\ref{app:results-centre}. Figure~\ref{fig:provenance} shows the mechanism on the largest instance. The archive's columns take a
sizable share of the master's pool, of the LP solution weight $\sum x$, and of the basis, so they do real work, and
the pricer replaces them only gradually. With the learned center the HEAVY iterations fall while
the demand for exact pricing does not, so its share of the run rises, which is what GPU pricing,
the third component, addresses in the next subsection.
\begin{figure}[!tb]\centering
\setlength{\abovecaptionskip}{2pt}\setlength{\belowcaptionskip}{-0.1pt}
\includegraphics[width=0.8\textwidth]{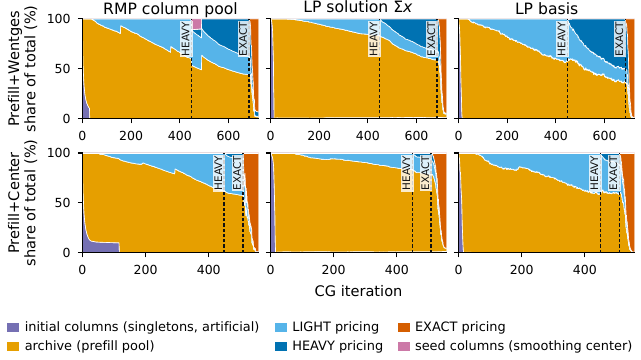}
\caption{Share of columns by column origin over the root's iterations on \provInst{}, for Prefill+Wentges (top) and Prefill+Center (bottom); dashed lines mark the pricer changes, LIGHT to HEAVY to EXACT.}
\label{fig:provenance}
\end{figure}

\subsection{GPU exact pricing}
\label{sec:exp-gpu}

\begin{wrapfigure}{r}{0.44\textwidth}
\vspace{-10pt}\centering
\includegraphics[width=0.44\textwidth]{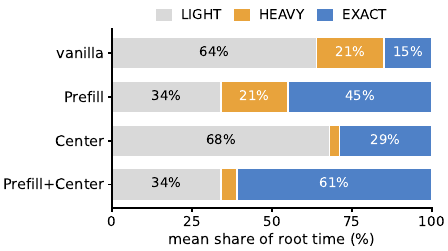}
\vspace{-6pt}\caption{Mean share of root time per pricing level on the panel.}
\label{fig:share}
\vspace{-8pt}
\end{wrapfigure}
Prefill and the center change where root pricing spends its time. Figure~\ref{fig:share} shows the
mean share of root time per pricing level on the panel: prefill mainly compresses the LIGHT level,
the center mainly compresses HEAVY, and together they raise the share of exact pricing from
\shareVanExact{}\,\% to \shareNcpuExact{}\,\%, so exact pricing becomes the largest stage of the run once
archive and center are in place, and it is where a further lever pays. Table~\ref{tab:gpu}
shows what this buys per exact call on \gpuHwLargestInst{}, the largest instance, measured as
pricing time alone without the LP solves. Over the ten X instances with the longest exact pricing
time (Appendix~\ref{app:gpu-percall}), the median ratio of CPU to GPU median call time is \gpuHwLFourOverCpuSpeedMed{} on a
low-power L4 and \gpuHwRtxOverCpuSpeedMed{} on the Blackwell card we deploy, with identical bounds on every GPU and identical exact-call counts
across the three cards. Because
exact pricing dominates the profile once prefill and center are in place, the per-call gain
converts into the full-stack result: the backend alone on vanilla RouteOpt gives a
root-time ratio of \ablVfourMainGGM{}; within the learned stack the GPU-to-CPU ratio is \ablVfourMainGpuOnStackPooledGM{} on the same \ablVfourMainGpuOnStackPooledN{} instances; and the full pipeline is \nhlRtxVfourMainSpeed{} times faster
than vanilla RouteOpt on average and \nhlRtxVfourMainBestSpeed{} times faster in the best case.

\subsection{Case study: parallel-machine scheduling}
\label{sec:exp-upms}

\input{sections/table_upms}

The scheduling oracle is a dense dynamic program, so this case study evaluates prefill and dual
smoothing without porting the routing GPU backend. We run inside the BPC solver of
\citet{hijazi2026improving}, on their instances and seeds, with a vectorized exact pricing oracle
shared by every arm so that the comparison is about column generation and not about the speed of
one implementation (Appendix~\ref{app:bpc-upms}); prefill here preloads the RMP with schedules from randomized list
scheduling, without the neural scorer, since the pricing subproblem of this problem is not hard.
Table~\ref{tab:upms} reports geometric-mean wall time with the number of instances certified within
the time limit. Prefill alone is already strong on this problem: it certifies the same instances as the baseline column generation in a small
fraction of the time. Wentges smoothing helps on its own, smoothing with prefill certifies all but one
instance, and adding the learned center certifies all \upmsAllN{} instances, where the learned
pricer certifies \upmsBodyHAllCert{}; the whole configuration is \upmsHOverHwPCAllSpeed{} times faster in geometric mean than the
baseline column generation. Changing how column generation is started and steered is therefore a lever of great power
on this problem, one that acts on the initialization and the duals rather than on the pricer.

%% file: sections/tables.tex
\begin{table}[!tb]
\centering
\scriptsize
\caption{Component ablation on the \ablVfourMainInstN{}-instance panel: geometric-mean root-time ratio to vanilla (lower is better), with the number of instances faster than vanilla, out of the cell's instances, in parentheses.}
\label{tab:ablation}
\setlength{\tabcolsep}{4pt}
\begin{tabular}{llrrrrrr}
\toprule
Set & Class & Inst. & Prefill & Center & GPU & Prefill+Center & Full pipeline \\
\midrule
\ablTableVfourMainRowsWords
\bottomrule
\end{tabular}
\end{table}

\begin{table}[!tb]
\centering
\footnotesize
\caption{Median seconds per steady-state exact call on \gpuHwLargestInst{} (learned stack), one CPU thread against three GPUs; same bound and call count on every card.}
\label{tab:gpu}
\setlength{\tabcolsep}{8pt}
\begin{tabular}{lcccc}
\toprule
 & CPU (one thread) & L4 & RTX PRO 6000 & B200 \\
\midrule
Seconds per exact call & \gpuHwLargestCpuSec{} & \gpuHwLargestLFourSec{} & \gpuHwLargestRtxSec{} & \gpuHwLargestBTwoSec{} \\
\bottomrule
\end{tabular}
\end{table}

%% file: sections/table_upms.tex
\begin{table}[!tb]
\centering
\footnotesize
\caption{Parallel-machine scheduling inside the code of \citet{hijazi2026improving}: geometric-mean wall time in seconds with the number of instances certified within $1800$\,s in parentheses; $m$ machines, $n$ jobs, five instances per cell; a star marks a cell with a timed-out run, which enters the mean at $1800$\,s. \emph{Wentges} is smoothing alone.}
\label{tab:upms}
\setlength{\tabcolsep}{4pt}
\begin{tabular}{lrrrrrr}
\toprule
 & $m4\,n40$ & $m4\,n100$ & $m8\,n40$ & $m12\,n100$ & $m20\,n100$ & all $25$ \\
\midrule
Their CG                        & \upmsBodyHMFourNForty{}   & \upmsBodyHMFourNHundred{}   & \upmsBodyHMEightNForty{}   & \upmsBodyHMTwelveNHundred{}   & \upmsBodyHMTwentyNHundred{}   & \upmsBodyHAll{} \\
Their transformer pricing               & \upmsBodyHnnMFourNForty{} & \upmsBodyHnnMFourNHundred{} & \upmsBodyHnnMEightNForty{} & \upmsBodyHnnMTwelveNHundred{} & \upmsBodyHnnMTwentyNHundred{} & \upmsBodyHnnAll{} \\
Prefill                         & \upmsBodyHPMFourNForty{}  & \upmsBodyHPMFourNHundred{}  & \upmsBodyHPMEightNForty{}  & \upmsBodyHPMTwelveNHundred{}  & \upmsBodyHPMTwentyNHundred{}  & \upmsBodyHPAll{} \\
Wentges                         & \upmsBodyHwMFourNForty{}  & \upmsBodyHwMFourNHundred{}  & \upmsBodyHwMEightNForty{}  & \upmsBodyHwMTwelveNHundred{}  & \upmsBodyHwMTwentyNHundred{}  & \upmsBodyHwAll{} \\
Wentges + Prefill               & \upmsBodyHwPMFourNForty{} & \upmsBodyHwPMFourNHundred{} & \upmsBodyHwPMEightNForty{} & \upmsBodyHwPMTwelveNHundred{} & \upmsBodyHwPMTwentyNHundred{} & \upmsBodyHwPAll{} \\
Wentges + Prefill + Center      & \upmsBodyHwPCMFourNForty{} & \upmsBodyHwPCMFourNHundred{} & \upmsBodyHwPCMEightNForty{} & \upmsBodyHwPCMTwelveNHundred{} & \upmsBodyHwPCMTwentyNHundred{} & \upmsBodyHwPCAll{} \\
\bottomrule
\end{tabular}
\end{table}

%% file: sections/conclusion.tex
\section{Conclusion}
\label{sec:conclusion}

This paper breaks with the folklore that initial columns harm column generation and turns
prefilling into a general plug-in accelerator for BPC: learning to prefill the
columns. The idea at its center is that imperfect columns, routes that need not belong to any optimal
solution but together cover the customers well, are what speeds the root up, provided they are held
outside the master as a lazy archive and served at the current smoothed dual. We built a
learn-to-prefill pipeline on this idea and obtained large speed-ups on two problems inside two exact
solvers, CVRP in RouteOpt and parallel-machine scheduling in the code of
\citet{hijazi2026improving}. On the hardest instances prefill also showed its weakness: the dual it
produces is close to the optimal dual in value but deviates from it in direction, so the query
needs a correction of direction rather than of distance, and we designed a learned dual center for
that purpose. Once archive and center are in place, the remaining bottleneck of the pipeline is
exact pricing, and we rebuilt it around its dominance test for the GPU. On the state-of-the-art CVRP
solver the complete pipeline reaches the certified root bound with a
\nhlRtxVfourMainSpeed{}-fold geometric-mean speed-up of root pricing over vanilla RouteOpt, and in
the best case in \nhlRtxVfourMainBestPct{}\,\% of vanilla RouteOpt's time.

The main limitation is that the effect on the full BPC tree is unknown: at the
scale we test, the instances have no known exact solution time, and root pricing is the part we can
measure and certify; extending the measurement to the tree is one direction of future work. A
second direction is the co-design of the two learned components, which we designed independently,
either as a joint configuration or as a single model that predicts pool and center together.